\pdfoutput=1
\documentclass[twoside,11pt]{article}
\usepackage[T1]{fontenc}
\usepackage{jmlr2e}
\usepackage{xcolor}
\usepackage{amsmath}
\usepackage{amsfonts}
\usepackage{amssymb}

\usepackage{lastpage}
\jmlrheading{XX}{202X}{1-\pageref{LastPage}}{XX/XX; Revised XX/XX}{XX/XX}{21-0000}{Jim Smith, Richard Preen, Andrew McCarthy, Maha Albashir, Alba Crespi-Boixader, Shahzad Mumtaz, Christian Cole, James Liley, Jost Migenda, Simon Rogers, and Yola Jones}

\ShortHeadings{SACRO-ML}{Smith~et~al.}
\firstpageno{1}

\begin{document}

\author{\name Jim Smith \email james.smith@uwe.ac.uk\\
    \name Richard Preen\\
    \name Andrew McCarthy\\
    \name Maha Albashir\\
    \addr School of Computer Science, University of the West of England
    \AND
    \name Alba Crespi-Boixader\\
    \name Shahzad Mumtaz\\
    \name Christian Cole\\
    \addr School of Medicine, University of Dundee
    \AND
    \name James Liley\\
    \addr Department of Mathematical Sciences, University of Durham
    \AND
    \name Jost Migenda\\
    \addr Department of Physics, King's College London
    \AND
    \name Simon Rogers\\
    \name Yola Jones\\
    \addr NHS National Services Scotland
}

\editor{}
\title{Safe machine learning model release from Trusted Research Environments: The SACRO-ML package}
\maketitle

\begin{abstract}
We present \texttt{SACRO-ML}, an integrated suite of open source Python tools to facilitate the statistical disclosure control (SDC) of machine learning (ML) models trained on confidential data prior to public release. \texttt{SACRO-ML} combines (i) a \textit{SafeModel} package that extends commonly used ML models to provide \textit{ante-hoc} SDC by assessing the vulnerability of disclosure posed by the training regime; and (ii) an \textit{Attacks} package that provides \textit{post-hoc} SDC by rigorously assessing the empirical disclosure risk of a model through a variety of simulated attacks \textit{after} training. The \texttt{SACRO-ML} code and documentation are available under an MIT license at \url{https://github.com/AI-SDC/SACRO-ML}.
\end{abstract}


\section[Introduction]{Introduction}

Trusted research environments (TREs) offer a means for researchers to analyse confidential datasets and publish their findings. The globally recognised `Five Safes' approach to privacy preservation and regulatory compliance~\citep{Green:2023} requires TRE staff to independently check the disclosure risk posed by any proposed output before release is approved. This process of statistical disclosure control (SDC) is well understood for outputs of traditional statistical analyses. However, while the use of machine learning (ML) methods has rapidly expanded, a recent study of TREs by \cite{TREsurvey} in the UK revealed a shortfall of SDC understanding and tools to support the use of ML. In particular, while researchers can already use ML methods within TREs, the inability to export those trained models precludes reproducibility of results, a critical tenet of scientific practice~\citep{begley15,arnold19}. After \cite{fredrikson2015model} demonstrated that ML models can be vulnerable to attacks inferring aspects of training data, there has been a rapid growth in proposed attacks and counter-measures as surveyed by \cite{hu2022membership}. Regulatory authorities such as the UK \cite{ico2022} recognise this risk, but offer general advice rather than practical guidance and tools. 

While several toolkits, such as \texttt{ML Privacy Meter}~\citep{mlprivacymeter}, \texttt{TensorFlow Privacy}~\citep{tfprivacy}, and \texttt{Adversarial Robustness Toolbox}~\citep{art}, offer help for researchers by providing various model specific privacy attacks, it is not clear that any of these individually meet the needs of TREs. Firstly, most are model-specific and typically require significant expertise to run and comprehend the metrics presented. Secondly, since the field is rapidly evolving, TREs need to be able to estimate an upper bound on metrics of \textit{future} risk, rather than only from attacks present at the time of release. Thirdly, a lack of understanding of the way TREs assess risk has lead to a focus within the ML-privacy community on metrics such as mean risk/AUC values. However, as \cite{ye2022enhanced} note: \textit{``Such an average-case formulation does not support reasoning about the privacy risk of individual data records''}---which forms the basis of `traditional' SDC when, for example, checking the disclosure risk of each cell in a table. Finally, TRE staff also routinely check for the unintended inclusion of data, etc., in tabular results. Hence, to avoid introducing additional delays in both the collection and processing of data, information about risk should be embedded within the research-SDC workflow, rather than requiring subsequent effort. 

A recent report by \cite{dare2022} highlighted the need to \textit{``Where possible, automate the review of outputs to support and focus the use of skilled personnel on the areas of most significant risk''}. Moreover, \cite{graimatter_report2022} have recently presented a detailed discussion of the implications of ML model release and proposed a set of guidelines for TREs. \texttt{SACRO-ML} makes the following contributions that aim to address these issues and enable ML use within TREs, fostering scientific discovery and aiding reproducibility of results:

\begin{itemize}
    \item A \textit{SafeModel} package that extends commonly used ML models to provide \textit{ante-hoc} SDC by assessing the theoretical risk posed by the training regime (such as hyperparameter, dataset, and architecture combinations) \textit{before} (potentially) costly model fitting is performed. In addition, it ensures that best practice is followed with respect to privacy, e.g., using differential privacy optimisers where available.
    \item An \textit{Attacks} package that provides \textit{post-hoc} SDC by rigorously assessing the empirical disclosure risk of a model through a variety of simulated attacks \textit{after} training. The package provides an integrated suite of attacks with a common application programming interface (API) and is designed to support the inclusion of additional state-of-the-art attacks as they become available. \texttt{SACRO-ML v1.2.1} provides implementations of membership inference attacks using the likelihood ratio (LiRA) approach described by \cite{carlini2022membership}, and worst-case estimation attacks for both membership and attribute inference. 
    \item Summaries of the results are written in a simple human-readable report. For large models and datasets, \textit{ante-hoc} analysis has the potential for significant time and cost savings by helping to avoid wasting resources training models that are likely to be found to be disclosive after running (potentially) intensive \textit{post-hoc} analysis. Examples of the risks addressed by \texttt{SACRO-ML} are provided since not all TRE staff and accredited researchers will be aware of the literature around ML model vulnerability.
\end{itemize}



\section{The \texttt{SACRO-ML} Package: An Overview}

\texttt{SACRO-ML} is available at \url{https://github.com/AI-SDC/SACRO-ML} under an MIT license. \texttt{SACRO-ML v1.2.1} supports Python 3.9--3.11 and utilises continuous integration tools to enforce a consistent style with \texttt{Ruff}~\citep{Ruff}, quality with \texttt{pylint}~\citep{pylint} and accuracy with \texttt{pytest}~\citep{pytest}. \texttt{Sphinx} is used to generate documentation deployed to GitHub Pages. The documentation provides an API reference as well as examples of how to create \textit{ante-hoc} checks for additional ML models via the \textit{SafeModel} package and integrate new \textit{post-hoc} privacy attacks within the common \textit{Attacks} API. Moreover, digital object identifiers are automatically assigned by \url{zenodo.org} for each release.

\subsection[SafeModels]{SafeModels}

The \textit{SafeModel} package provides a set of privacy-enhancing wrapper classes designed with an object oriented approach that enables an extensible API using multiple inheritance to easily add new ML models. Currently implemented are a number of standard \texttt{sklearn}~\citep{scikitlearn} models, including the Decision Tree Classifier (DT), Random Forest Classifier (RF), and Support Vector Classifier (SVC). Also supported is TensorFlow's \texttt{keras} model~\citep{keras}. Using multiple inheritance in this way allows researchers to use standard libraries with no changes to their normal workflow, while helping them adopt best-practice and enabling automated assistance to output checkers.

The \textit{SafeModel} superclass extends existing functionality such that: (i) Researchers are warned when they have chosen hyperparameters (often the default values) that are likely to result in disclosure, e.g., by overfitting. (ii) Differentially private~\citep[DP;][]{dwork08} algorithms are automatically called by \texttt {compile()} and \texttt{fit()}. (iii) Researchers may query model safety via methods such as \texttt{check\_epsilon()} for DP, and \texttt{run\_attack()}. (iv) Poor practice, such as changing hyperparameters without retraining the model, is automatically detected and \textit{post-hoc} privacy attacks are run to assess model disclosure. Detailed JSON-formatted and PDF reports are produced by \texttt{request\_release()} to assist TRE output checkers. Our design philosophy is that researchers and TRE checkers should have sufficient training to know that \textit{some} hyperparameter settings are likely to create disclosive models. However, while legally responsible for their choices, they cannot be expected to remember \textit{which} are disclosive. Hence, \textit{SafeModel} takes values from a read-only JSON file where the TRE defines their risk appetite. For example, minimum values for $\epsilon$ (SVC/Keras) and \texttt{min\_samples\_leaf} (DT/RF) where defaults (1 in the latter case) are risky. As the field generates new understanding, these can be simply updated by the TRE staff.


\subsection[Attacks]{Attacks}

Our simulated attacks work with any target model that extends the \texttt{BaseEstimator} from \texttt{sklearn}, including, for example, models that are not part of the \texttt{sklearn} package such as the \texttt{XGBClassifier} model within XGBoost. Attacks can also be run in a model agnostic way, based upon saved model predictions and could therefore be used to assess models developed in languages other than Python. Two key membership inference attacks are implemented and the hierarchical class structure used makes it possible to extend to future attacks as they are developed and published. \texttt{SACRO-ML v1.2.1} uses Random Forest as the default attack model, but any others can be used.

The LiRA~\citep{carlini2022membership} membership inference attack is provided as representative of the current state-of-the-art, and because it rightly places an emphasis on an attacker scenario that corresponds to TRE concerns: \textit{``If a membership inference attack can reliably violate the privacy of even just a few users in a sensitive dataset, it has succeeded''}. 

Once a model is released from a TRE, researchers and staff cannot control what new attacks may be used against it. Hence \texttt{SACRO-ML} also implements methods that aim to provide TREs with an assessment of the worst-case vulnerability. The worst-case membership inference attack trains models using the target model's outputs for the same train and test splits provided to the researcher, thus removing the uncertainty associated with both model and data. This equates to an even stronger upper-bound than the `(Idealized) Attack L: Leave-one-out attack'~\citep{ye2022enhanced}.

The attribute inference attack also implements a worst-case attack. It does this by assuming the attacker has access to a record missing the value for one attribute, and measuring whether the trained model allows more (and more accurate) predicted completions for the training set than it does for the test set. An exhaustive search of the target model's predictive confidence is performed for all possible values that complete the record (discretised for continuous attributes). The attack model then makes a prediction if one missing value (categorical) or a single unbroken range of values (continuous) leads to the highest target confidence, which is above a user-defined threshold; otherwise it reports \textit{don't know}.

The attack computes an upper bound on the fraction of records that are vulnerable, i.e., where the attack makes a correct prediction, and reports the Attribute Risk Ratio $ARR(a)$: the ratio of training and test set proportions for each attribute $a$. The attack is considered accurate if the target model's predicted label $l^*$ for the record with a single missing value is the same as for the actual record value $l$ (categorical) or the range of values yielding the same target confidence lies within $l\pm10\%$ (continuous). This latter condition mirrors the protection limits commonly used in cell suppression algorithms; see, for example, \cite{smith2012genetic}. The $ARR$ metric recognises that any useful trained model contains some generalisable information and so only considers the model to be leaking privacy if $ARR(a)>1$. It also recognises that not all attributes will be considered equally disclosive, so enables a discussion between TRE staff and researchers. 

\section[Conclusions]{Conclusion and Future Work}

\texttt{SACRO-ML} is the first framework to combine a range of privacy attacks with rules-based \textit{ante-hoc} SDC for different models under a common API, and with standard outputs for interpretability and interoperability. It serves as a focal point for researchers to learn about ML privacy, and for TREs to assess the risks of releasing trained models into the wild. We welcome community engagement and there is an ongoing commitment to extend the range of models and attacks supported, and improve interpretability of outputs. Future releases will include a wider range of attacks and incorporate additional open source privacy assessment tools for \textit{post-hoc} SDC. This will allow researchers to measure the `reality gap' between worst-case risk and estimates provided by state-of-the-art algorithms such as LiRA.

\bibliography{references}

\section*{Acknowledgement}

This work was funded by UK Research and Innovation under Grant Numbers MC\_PC\_21033 and MC\_PC\_23006 as part of Phase 1 of the Data and Analytics Research Environments UK (DARE UK) programme, delivered in partnership with Health Data Research UK (HDR UK) and Administrative Data Research UK (ADR UK). The specific projects were Semi-Automatic checking of Research Outputs (SACRO; MC\_PC\_23006) and Guidelines and Resources for AI Model Access from TrusTEd Research environments (GRAIMATTER; MC\_PC\_21033). It has also been supported by MRC and EPSRC (PICTURES; MR/S010351/1).

\end{document}